%% file: architecture_arxiv.tex
\documentclass[a4paper]{article}

\usepackage[includeheadfoot, left=3cm, right=3cm, top=2cm, bottom = 2cm]{geometry}
\usepackage{fancyhdr}
\usepackage{authblk}

\usepackage[xindy]{glossaries}
\loadglsentries{acronyms}

\usepackage[utf8]{inputenc}
\usepackage[T1]{fontenc}
\usepackage{libertine}
\usepackage[UKenglish]{isodate}

\usepackage{amsmath}
\usepackage{graphicx}
\usepackage{xcolor}
\usepackage{textcomp}

\usepackage{cite}

\usepackage[hidelinks]{hyperref}
\usepackage{doi}
\usepackage[strings]{underscore}
\usepackage{orcidlink}

\makeglossaries

\usepackage[
    type={CC},
    modifier={by},
    version={4.0},
]{doclicense}

\fancypagestyle{firststyle}
{  
   \fancyhf{}
   \fancyfoot[L]{
   \begin{minipage}{\textwidth}
   \centering {\footnotesize \doclicenseThis }
   \thepage 
\end{minipage}  
}
 
}
\begin{document}

\title{An Abstract Architecture for Explainable Autonomy in Hazardous Environments}

\author[1]{Matt Luckcuck \orcidlink{0000-0002-6444-9312}}
\author[2]{Hazel M Taylor \orcidlink{0000-0003-0130-1174} }
\author[1]{Marie Farrell \orcidlink{0000-0001-7708-3877} }

\date{ 20\textsuperscript{th} of October 2022}

\affil[1]{Department of Computer Science \\ \& Hamilton Institute,\\ Maynooth University,\\ Ireland}
\affil[2]{Department of Computer Science,\\ University of Manchester,\\ United Kingdom}

\maketitle 
\thispagestyle{firststyle}

\begin{abstract}
Autonomous robotic systems are being proposed for use in hazardous environments, often to reduce the risks to 
human workers. In the immediate future, it is likely that human workers will continue to use and direct these autonomous robots, much like other computerised tools but with more sophisticated decision-making. Therefore, one important area on which to focus engineering effort is ensuring that these users \textit{trust} the system.
Recent literature suggests that \textit{explainability} is closely related to how trustworthy a system is. Like safety and security properties, explainability should be designed into a system, instead of being added afterwards. 
This paper presents an abstract architecture that supports an autonomous system explaining its behaviour (explainable autonomy), providing a design template for implementing explainable autonomous systems. 
We present a worked example of how our architecture could be applied in the civil nuclear industry, where both workers and regulators need to trust the system's decision-making capabilities.
\end{abstract}

\section{Introduction}

Autonomous robotic systems (robotic systems  controlled by software capable of making decisions without human intervention) are being used, or proposed for use, in a variety of environments that are hazardous to humans. For the nuclear industry, these hazards are often chemical or radiation; for other hazardous environments, they could include: extreme weather, high pressure, or lack of breathable atmosphere~\cite{fisher_overview_2021}. 
Using autonomous robots in these environments reduces the risk of harm to humans, or enables tasks that humans are not capable of. 
These tasks could include: remote inspections, e.g.~\cite{cardoso2020heterogeneous, cardoso_towards_2020}; remote handling, or laser cutting~\cite{luckcuck_workshop_2020}.
However, introducing autonomy to sectors like the nuclear industry (which are often, needfully, cautious about new technology) requires that the systems are trusted; trusted by the workers using them and by the sector's regulator. 

Our previous work presents high-level principles for autonomous robotics systems that will be used in hazardous environments~\cite{luckcuck_principles_2021}, which was co-authored by representatives from the Office for Nuclear Regulation, the UK's civilian nuclear regulator. In that work, we argue that a system's autonomous components ``should be as transparent and verifiable as possible'', and 
``demonstrably trustworthy'' to operators.

Recent work has shown that the concept of \textit{explainability} supports a system being trustworthy.  
Balasubramaniam et al.~\cite{balasubramaniam_transparency_2022} conducted a study showing that ``explainability is tightly coupled to transparency and trustworthiness of AI systems.'' Loss of human trust in a system is one of the \textit{ethical hazards} identified in the `Ethical Design and Application of Robots and Robotic Systems' standard~\cite{EthicalDesignofRobots2016}, which suggests using explanations to help restore user trust.
Explanations can also compensate for  mildly undesirable agent behaviour~\cite{stange_effects_2020}. 
Similarly to safety, security, or other ethical hazards identified in~\cite{EthicalDesignofRobots2016}, explainability should be designed into the system rather than hoping that it can be added during development~\cite{luckcuck_matt_principles_2021}.

This vision paper describes an abstract architecture for building explainable autonomous systems: autonomous systems that can explain and justify their decisions to humans.
We refer to this as \textit{explainable autonomy}, to indicate that both symbolic and sub-symbolic AI techniques can be included. 
Our architecture separates high-level and low-level decision-making. We propose that a \gls{bdi}~\cite{bratman:87a, rao:95b} agent is used the make the high-level decisions, which is a style of symbolic AI, where an agent interprets information about its environment into logical statements, which it then uses to rationally choose which of a set of plans to perform. An agent's \textit{beliefs} are its knowledge about itself and its environment, its \textit{desires} are the agent's long-term goals, and its \textit{intentions} are the goals it is currently pursuing. Using \gls{bdi} enables executive decisions to be verified~\cite[Recipe 1]{Luckcuck_using_2021}~\cite{luckcuck_principles_2021}.

Our architecture supports explanations in two ways. 
1) the \textit{Central Executive} (the \gls{bdi} Agent and Explainer) receives all of the system's information \textit{and} makes its executive decisions, making it easier to build explanations. 
This is similar to the proposal of an Ethical Black Box~\cite{winfield_jirotka_ehticalBalckBox}.
2) the \gls{bdi} Agent reasons logically, using knowledge and beliefs, which supports building an explanation.
This enables two modes of explanation. 
1) explanations about previous actions, e.g. ``Why did you take this route instead of that route?'', which justify the system's behaviour.
2) explanations about future action/inaction, e.g. ``Why \textit{can't} you move over there?'', which check that the system will obey a safety rule and give a deeper insight into how the system works, thereby enhancing safety.

\S~\ref{sec:explanations} overviews eliciting a system's explainability requirements, though this is not the paper's main focus. \S~\ref{sec:engineering} describes our architecture, its Central Executive, and the verification benefits of our approach. \S~\ref{sec:example} presents a worked example of how our architecture supports explanations. Finally, \S~\ref{sec:conclusion} gives our concluding remarks.

\section{Designing Explainability}
\label{sec:explanations}

A system's \textit{explainability requirements} -- what and how it needs to provide explanations -- need to be carefully elicited, because there is no easily reusable detailed definition of what it means for an autonomous system to be explainable; or how this concept overlaps with transparency, explicability, or traceability. 
A useful explanation should be easily understood by humans, but (in contrast to other software) autonomous systems may make surprising decisions, which means that useful explanations are not always obvious.
To design effective explainability, one must know the system’s stakeholders and context: to whom are you explaining, and what the system will be doing~\cite{brunotte_quo_2022}.

Our work focusses on explaining past action and future inaction; both of which will be requested by a human asking ``Why'' questions~\cite{gilpin_explaining_2018}.
Explaining past action is useful to check the decision-making behind (either correct or incorrect) behaviour. This is likely, especially in social settings, to be requested by contrastive questions, e.g: ``Why did you do \textit{that} instead of \textit{this}?''~\cite{Miller2019}.
Explanations of future inaction answer queries about barriers to performing behaviour. These are requested by questions of the form: ``Why can’t you do \textit{this}?'', similar to the system in~\cite{Winikoff_debugging_2017}. 
Our position is that supporting two types of explanation can help a wide range of stakeholders -- including users and regulators -- to trust the decisions that the system has made, and that it might make in the future.

The system's explainability requirements should be carefully elicited, using representative samples of \textbf{all} of the system's stakeholders, and considering the system's context. 
The explanations should suit all stakeholders who will depend on explanations for their trust in the system, which means that the types of stakeholders should be carefully identified, e.g. using Tomsett et al.'s model of stakeholder roles~\cite{Tomsett2018}.
Considering the context aims to produce explanations that are appropriate for the system's operating environment.

Safety-critical systems are likely to have additional explanation requirements~\cite{hall2019systematic}.
For example, in an emergency situation where the system is being asked to explain its behaviour, it may be appropriate to provide an explanation with less detail, to avoid wasting time that could be used to avert a dangerous situation~\cite{brunotte_quo_2022}. Even in nominal situations, the level of detail in an explanation should not overload operators with information, which could delay their own decisions at a critical moment.

There is emerging research on how best to provide explanations. 
For example, causal explanations that refer to a behaviour's intended goal are often preferred~\cite{Broekens_user_evaluated, stange_effects_2020}, but the explanation of some behaviours benefit from referring to its enabling conditions (guards)~\cite{Broekens_user_evaluated}.
However, these were both small user studies (n=30~\cite{Broekens_user_evaluated} and n=38~\cite{stange_effects_2020}), and the authors of~\cite{stange_effects_2020} suggest their result could be due to the causal explanations being longer and more detailed than the others that they tested.
Also, both of these studies build the system and then test it with users. However, stakeholders (especially end-users or operators) should be involved in the design and evaluation of the explainability; this is important in general, as well as in the nuclear domain.

Approaches for including stakeholders in requirements elicitation are being developed.
In previous work, we begin to interpret the factors of an explanation that stakeholders need in safety-critical nuclear environments 
by presenting stakeholders (n=16) with scenarios, asking what information they want to know, and comparing the findings to how people explain~\cite{DOI}.
A similar approach prompts users with vignettes involving ethical requirements, rather than asking about them directly, as a way of side-stepping social desirability bias~\cite{Negri-Ribalta_fse}.
Another approach combines user-centred and participatory design for robotic systems~\cite{Winkl2020}, yielding higher participant system acceptance (than before their study). The case study was a socially assistive robot, but given that explanations are a social interaction~\cite{Miller2019}, their approach is likely to be applicable in eliciting explainability requirements.

Requirements elicitation approaches from the literature should be used 
to ensure that the system's explainability requirements are valid, unbiased, and unambiguous. Our focus on past and future (in)actions is supported by the literature, but may need extending to meet stakeholder requirements for a specific system or context. The literature on requirements elicitation is moving towards more human-centred approaches to designing explanations (e.g.~\cite{Wolf2019, Cirqueira2020}) but this is not yet standard practice. The next section introduces our architecture, which provides a target for implementing explainable autonomy. We intend this to support and promote further research into requirements elicitation for explainable autonomy.

\section{An Explainable Autonomy Architecture }
\label{sec:engineering}

In this section, we describe our abstract explainable autonomy architecture. 
Following our suggestion in prior work~\cite{Luckcuck_using_2021}, 
our architecture (Fig.~\ref{fig:arch}) uses a rational \gls{bdi} agent to make executive decisions, e.g. choosing the next waypoint or scheduling tasks. 
The symbolic AI techniques that underpin \gls{bdi} agents are explicit and based on mathematical logic, so they are (relatively) easy to inspect. 
We restrict connectionist/sub-symbolic AI (such as Machine Learning or Neural Networks) to more granular functions, e.g. vision classification. 

Existing work in the literature explores the utility of \gls{bdi} agents in providing explanations. For example, by extracting a chain of trace events (stored internally by the agent) that go back through the system's runtime to explain behaviour~\cite{koeman_why_2020}. 

This decision-making Agent and the Explainer component (which builds explanations) form the \textit{Central Executive}
\footnote{We borrow the term \textit{Central Executive} from Baddely and Hitch's Working Memory Model~\cite{Baddely_Hitch_1974}.} 
Our architecture routes all of the system's information to this Central Executive, enabling it to collect and store the information needed to produce an explanation.
\textit{What} information is needed for explanations is described by the system's explainability requirements (\S~\ref{sec:explanations}), but this routing ensures that all of the information is available to the Explainer.
Clearly, if a part of the system operates entirely separately from the components that make the decisions and  produce the explanations, it will be difficult to explain its behaviour.

\subsection{High-Level Architecture}
\label{sec:arch}

\begin{figure}
\centering
\includegraphics[width=0.47\textwidth]{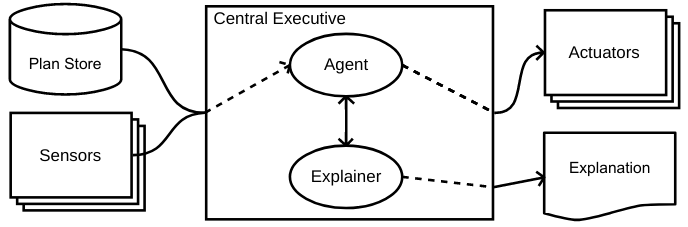}
\caption{High-Level view of our abstract architecture. The squares are system modules (which may contain hardware and software), the cylinder is a data store, and the square with the curved base is the explanation `document' (which could be text, speech, etc.). The ovals are sub-modules in the Central Executive. The arrows show the information flow, with the dashed lines for inside the Central Executive. 
 \label{fig:arch}}
\end{figure}

Fig.~\ref{fig:arch} gives a high-level view of our architecture: an abstracted autonomous robotic system that has sensors, actuators, and a Central Executive. 
The sensors and actuators enable the system to sense and interact with its environment;
both components are likely to be wrapped in software that interprets the raw sensor information or translates commands into low-level actuator control signals, respectively.
The system will likely have or build a model of its environment, but this is not shown in Fig.~\ref{fig:arch} because it is not the focus of our architecture.

Our architecture's key feature is that it routes all of the system's information through the Central Executive. This ensures that when an explanation is requested, the required information is available. To avoid bottlenecks, components may directly communicate as long as the Central Executive simultaneously receives the same information.

Here we outline the architecture, but we see several implementation routes. Robotic systems are often implemented using node-based software frameworks~\cite{Luckcuck2019}, the most well-used in the literature is the \gls{ros}~\cite{quigley2009ros}, and others include: \Genom{}\footnote{\Genom{}: \url{https://git.openrobots.org/projects/genom3}} and NASA's Core Flight System\footnote{Core Flight System: \url{https://cfs.gsfc.nasa.gov/}}. 
These frameworks often contain common concepts~\cite{Shakhimardanov2010}, which can enable reuse between frameworks~\cite{Ahn2017}. The modularity in our abstract architecture means that it is framework-agnostic.

\subsection{Central Executive}
\label{sec:ce}

As previously mentioned, the Central Executive comprises the \gls{bdi} Agent and an Explainer. These components collaborate to produce explanations.

The \gls{bdi} Agent makes the system's executive decisions; it decides on the system's behaviour, based on sensor input and the user-defined plans. The types of sensors needed will depend on the system's function and context. The raw data from the sensor hardware is likely to be processed by a software module before being sent to the Agent, which will interpret the information into logical properties that it can reason about (its beliefs). Its other input is a set of plans, written \textit{a priori} for the system's particular mission and environment\footnote{
Our approach should also be applicable to systems that learn new plans, as long as the Agent and Explainer both have access to the plan store.}. The Agent chooses a plan, based on its beliefs about the world and other information that it has stored, following it until new information triggers a different plan. The information about which events trigger which plans is used by the Explainer.

The Explainer produces explanations, in response to questions from an operator. The explanation's format (e.g. textual, graphical, vocal, etc.) is not restricted by our architecture. Both of these features depend on the system's explainability requirements (\S~\ref{sec:explanations}).
The Explainer stores information from the rest of the system (e.g. the raw sensor information and corresponding beliefs, which plan was triggered and why, etc.) to enable it to explain the system's behaviour.
The context of the question may also change the explanation that is produced. 
For example: the stakeholder's role (e.g. from~\cite{Tomsett2018}) may alter the level of detail in the explanation or its format; time-pressure on the reply may prompt a terser answer; or the question being about past action or future (in)action.

Work in the literature shows that a \gls{bdi} approach can be used to provide explanations of an agent's behaviour. For example, a log of a \gls{bdi} agent's goals and behaviour can be used to produce an explanation~\cite{Broekens_user_evaluated}.
Further, the work in~\cite{Kindricks_debugging_2012}, which highlights the utility of logic-based autonomy for providing explanations; and~\cite{Winikoff_debugging_2017}, which presents an approach to answer \textit{``why?''} and \textit{``why not?''} questions. However, both of these approaches work through debugging features, suggesting that their explanations may be mainly aimed at software developers. While this is undoubtedly useful, the explanations provided may not be suitable for other stakeholders.

When explaining past action, the Explainer can passively collect the information stored about the system and the Agent's decisions. However, when explaining future inaction, the Explainer needs to be able to trigger the Agent's reasoning cycle to evaluate a particular goal's feasibility. 
This `dry-run' functionality will likely require modifications to an agent to enable the Explainer to identify which plan corresponds to the requested action, and trigger the Agent's reasoning logic to investigate what is preventing that plan from being selected. 
\S~\ref{sec:example} gives a more detailed example of this functionality.

\subsection{Verification Benefits}
\label{sec:knockons}

The choice of a \gls{bdi} Agent enables the verification of the system's executive decisions, as recommended in our previous work~\cite[Recipe 1]{Luckcuck_using_2021}. In particular, \gls{bdi} agents are amenable to Formal Methods (FM), mathematically-based techniques for the specification and verification of systems, which can provide strong guarantees about software correctness.

For example, the \Gwen{} agent language~\cite{dennis2008gwendolen} can be exhaustively verified by its \textit{program model checker} the \gls{ajpf}~\cite{dennis2018mcapl}. A \Gwen{} program can be checked to prove that its decisions preserve formal properties, written in a temporal logic that has been extended with modalities capturing \gls{bdi} concepts such as \textit{the agent believes\ldots} so that its reasoning can be checked. The benefit of this approach is that \gls{ajpf} checks the program code that the agent will run during operation\footnote{\Gwen{} executes as ByteCode on a Java Virtual Machine. Nevertheless, \gls{ajpf} exhaustively verifies the agent's source code.}; but does so off-line, which does not impact the agent's performance. For an overview of the available FM approaches for autonomous systems see~\cite{Luckcuck2019}.

An example of a property could be that the agent will `eventually believe that it has reached the goal location', and \gls{ajpf} can check if this is true for the agent's program. 
These temporal logic properties could be translated into natural language~\cite{cherukuri_towards_2022}, to explain them to stakeholders without a formal background. This would help stakeholders to check that the verified properties correspond to the system's requirements.

\section{Explanation Examples}
\label{sec:example}
 
Here we present a worked example of how our abstract architecture (\S~\ref{sec:engineering}) supports explanations of both past and future (in)action. 
We consider a wheeled-rover using our architecture and Central Executive (\S~\ref{sec:arch}), performing a remote inspection mission in a nuclear waste store (adapted from~\cite[\S4.2.2]{robotics10030086}).  The rover's mission is to patrol waypoints inside a nuclear waste store, where low-level radioactive waste is stored in steel drums. The rover: takes radiation readings at each waypoint, to assess the integrity of the storage drums; maintains a given distance from the drums; and must avoid high-radiation areas if it can. The radiation readings and high-radiation areas are reported for later investigation.

To navigate around the waste store, the Agent decides on its plan based on the sensor information and its goal (to take radiation readings at each of the waypoints). As previously mentioned, all of the sensor information is routed through the Central Executive; which makes decisions, sends instructions to the actuators, and collects the information needed by the Explainer component. The explanations are realisable because \gls{bdi} logics capture the Agent's beliefs about the world. The Explainer has access to this information, facilitating explanations. Here, we describe how an explanation about past action and future inaction could be realised.

\subsection{Past Action}
\label{sec:pastActionExample}

In this example, the rover is navigating from Waypoint A to Waypoint B and the Agent interprets sensor information as an obstacle. Because the rover should not collide with obstacles, this new sensor information causes it to abort navigating to Waypoint B and navigate to Waypoint C instead. 

The rover's operator might ask ``why did you go to Waypoint~C instead of Waypoint B?''. Because the Central Executive has: received the sensor information that suggests there is an obstacle \textit{en route} to Waypoint B, interpreted the sensor information into a belief that there is an obstacle, and decided to change the plan; it has the information needed to answer the question. It could explain that ``my sensors tell me there is an obstacle ahead of me, between Waypoint A and Waypoint B; and I am not supposed to collide with obstacles.'' Here, different forms of explanation could be useful, e.g. the system could explain textually/verbally and provide an image of the perceived obstacle for the operator to check. As previously mentioned, work like~\cite{koeman_why_2020} provides a route to realise these explanations through the Agent's trace of events.

\subsection{Future Inaction}
\label{sec:futureInactionExample}

In this example, a nuclear inspector is checking that the rover's agent will not \textit{choose} to break safety rules, similarly to the expectations of a human worker. Here, the Central Executive could `dry-run' an instruction that should fail, and explain why it cannot be carried out.

For example, the inspector could ask ``why can't you go to Waypoint X?'', where Waypoint X is too close to a storage drum. Here, the Explainer would request the Agent to evaluate the instruction `go to Waypoint X' and interpret the result in much the same way as in \S~\ref{sec:pastActionExample}. As previously mentioned, this requires an Agent with a reasoning cycle modified so that it can be triggered to evaluate a particular instruction.

The Explainer will send the instruction to `go to Waypoint X' to the Agent, triggering it to attempt to enact the relevant plan. When the plan cannot be executed, the feedback from the Agent allows the Explainer to answer the inspector: ``I cannot go to Waypoint X because my map tells me that it closer to a storage drum than I am allowed to go''. 

This feature enables safety tests of the Agent's adherence to safety rules. It would also be extremely useful when testing the autonomous system before it was attached to the robotic system. A simulated environment could be used, and the safety properties could each be queried to provide confidence that the integrated autonomous robotic system will not be dangerous if used in field tests.
This ability to `dry-run' actions has appeared in the literature before, e.g. in~\cite{Dennis_ethical2015}, where they replicate the agent's reasoning internally to enable it to assess the ethical consequences of its own actions.

\section{Conclusion and Future Work}
\label{sec:conclusion}

This paper presents an abstract architecture for enabling explainable autonomy. The architecture routes the system's information through a Central Executive, which comprises a \gls{bdi} Agent and an Explainer component. We provide worked examples of how this approach supports explanation of past and future (in)actions.

The major area of future work is to develop and implement our architecture. 
We intend to use \gls{aadl} (or a similar notation) to develop a high-level design that is applicable to a variety of robotic software frameworks, exploiting their commonalities~\cite{Shakhimardanov2010}. Implementing the design will then enable us to demonstrate the architecture's utility.

Another important line of future work is to incorporate approaches that elicit explanation requirements, and assess their suitability for particular stakeholders. The approach in~\cite{Winkl2020} provides a useful starting point for involving stakeholders in the design process. The basis of our ongoing requirement elicitation work will be built on our previous study~\cite{DOI}, which involves eliciting requirements from nuclear sector stakeholders.

\bibliographystyle{plain}
\bibliography{roadmap-refs.bib}

\end{document}

%% file: acronyms.tex



\newcommand{\Genom}{G\textsuperscript{en}oM}

\newcommand{\Gwen}{\textsc{Gwendolen}}

\newacronym{ltl}{LTL}{Linear-time Temporal Logic}

\newacronym{foltl}{FOLTL}{First-Order Linear-time Temporal Logic}

\newacronym{ctl}{CTL}{Computation Tree Logic}

\newacronym{ptl}{PTL}{Probabilistic Temporal Logic}

\newacronym{pctl}{PCTL}{Probabilistic Computation Tree Logic}

\newacronym{pctl*}{PCTL*}{Probabilistic Computation Tree Logic*}

\newacronym{tctl}{TCTL}{Timed Computation Tree Logic}

\newacronym{stl}{STL}{Signal Temporal Logic}

\newacronym{ltl-x}{LTL-X}{a version of Linear Time Logic without the `next' operator}

\newacronym{mtl}{MTL}{Metric Temporal Logic}

\newacronym{iactlk-x}{IACTLK-X}{a fragment of the temporal-epistemic logic CTLK, without the `next' operator}

\newacronym{mucalc}{$\mu$-calculus}{a strict superset of other temporal logics}

\newacronym{fotl}{FOTL}{First-Order Temporal Logic}

\newacronym{bdi}{BDI}{Belief-Desire-Intention}

\newacronym{prs}{PRS}{Procedural Reasoning System}

\newacronym{are}{ARE}{Autonomous Reasoning Engine}

\newacronym{dmars}{dMARS}{distributed Multi-Agent Reasoning System}

\newacronym{mas}{MAS}{Multi-Agent System}


\newacronym{asm}{ASM}{Abstract State Machines}

\newacronym{fsm}{FSM}{Finite-State Machine}

\newacronym{pfsm}{PFSM}{Probabilistic Finite-State Machine}

\newacronym{apfsm}{APFSM}{\textit{Autonomous} Probabilistic Finite-State Machine}

\newacronym{pha}{PHA}{Probabilistic Hybrid Automata}

\newacronym{fsa}{FSA}{Finite-State Automata}

\newacronym{ta}{TA}{Timed Automata}

\newacronym{gspn}{GSPN}{Generalised Stochastic Petri Net}

\newacronym{mopn}{MOPN}{Marked Ordinary Petri Net}

\newacronym{ba}{BA}{B\"{u}chi Automata}

\newacronym{dtmc}{DTMC}{Discrete-Time Markov Chain}

\newacronym{ctmc}{CTMC}{Continuous-Time Markov Chain}


\newacronym{pars}{PARS}{Process Algebra for Robot Schemas}

\newacronym{fsp}{FSP}{Finite State Processes}

\newacronym{piadl}{$\pi$ADL}{$\pi$ calculus combined with ADL}

\newacronym{csp}{CSP}{Communicating Sequential Processes}

\newacronym{hcsp}{HCSP}{Hybrid Communicating Sequential Processes}

\newacronym{shcsp}{SHCSP}{Stochastic Hybrid Communicating Sequential Processes}

\newacronym{ccs}{CCS}{Calculus of Communicating Systems}

\newacronym{wsccs}{WSCCS}{Weighted Synchronous CCS}

\newacronym{csp|b}{CSP$\|$B}{an integrated formalism combining CSP and B}


\newacronym{vhdl}{VHDL}{VHSIC Hardware Description Language}

\newacronym{adl}{ADL}{Architecture Description Language}


\newacronym{fdr}{FDR}{the Failures-Divergences Refinement checker}

\newacronym{jpf}{JPF}{Java PathFinder}

\newacronym{ajpf}{AJPF}{Agent Java PathFinder}

\newacronym{mcmas}{MCMAS}{Model Checker for Multi-Agent Systems}

\newacronym{ltlmop}{LTLMoP}{Linear Temporal Logic MissiOn Planning}

\newacronym{cbmc}{CBMC}{C Bounded Model Checker}


\newacronym{psl}{PSL}{Property Specification Language}

\newacronym{fpga}{FPGA}{Field-Programmable Gate Array}

\newacronym{aadl}{AADL}{Architecture Analysis and Design Language}

\newacronym{lts}{LTS}{Labelled-Transition System}

\newacronym{uml}{UML}{Unified Modelling Language}

\newacronym{sysml}{SySML}{Systems Modelling Language}

\newacronym{robotml}{RobotML}{Robot Modelling Language}

\newacronym{dsl}{DSL}{Domain Specific Language}

\newacronym{sct}{SCT}{Supervisory Control Theory}

\newacronym{sat}{SAT}{Boolean satisfiability problem}

\newacronym{smt}{SMT}{Satisfiability Modulo Theory}

\newacronym{ide}{IDE}{Integrated Development Environment}

\newacronym{mde}{MDE}{Model-Driven Engineering}

\newacronym{ai}{AI}{Artificial Intelligence}

\newacronym{fdir}{FDIR}{Fault Detection, Isolation and Recovery}

\newacronym{ifm}{iFM}{Integrated Formal Methods}

\newacronym{ros}{ROS}{Robot Operating System}

\newacronym{amcl}{AMCL}{Adaptive Monte Carlo Localisation}

\newacronym{dl}{\text{\upshape\textsf{d{\kern-0.05em}L}}}{differential dynamic logic}

\newacronym{vm}{VM}{Virtual Machine}

\newacronym{bip}{BIP}{Behaviour Interaction Priority}

\newacronym{fol}{FOL}{First-Order Logic}

\newacronym{owl}{OWL}{Web Ontology Language}

\newacronym{ria}{RIA}{Restore Invariant Approach}

\newacronym{ocl}{OCL}{Object Constraint Language}

\newacronym{sil}{SIL}{Safety Integrity Level}

\newacronym{rcl}{RCL}{ROS Contract Language}

\newacronym{rv}{RV}{Runtime Verification}

\newacronym{rml}{RML}{Runtime Monitoring Language}

\newacronym{sats}{SATS}{Small Aircraft Transportation System}

\newacronym{ico}{ICO}{Interactive Cooperative Objects}

\newacronym{gui}{GUI}{Graphical User Interface}

\newacronym{can}{CAN}{Controller Area Network}

 
\newacronym{mtbf}{MTBF}{Mean Time Between Failures}

\newacronym{wcet}{WCET}{Worst-Case Execution Time}


\newacronym{caa}{CAA}{Civilian Aviation Authority}

\newacronym{atc}{ATC}{Air Traffic Control}

\newacronym{dlr}{DLR}{German Aerospace Centre}